\pdfoutput=1

\documentclass[11pt]{article}
\usepackage{graphicx} 
\usepackage{amsfonts}
\usepackage[final]{ACL2023}

\usepackage{amsmath}
\usepackage{amssymb}

\DeclareMathOperator*{\argmax}{arg\,max}

\usepackage{xcolor}
\usepackage{colortbl}
\usepackage{caption}
\usepackage{hyperref}
\usepackage{url}
\usepackage{graphicx}
\usepackage{booktabs}
\usepackage{multirow}
\usepackage{cleveref}
\usepackage{array}
\usepackage{listings}
\usepackage{colortbl}
\usepackage{caption}
\usepackage{enumitem}
\usepackage{wrapfig}
\usepackage{subfigure}
\usepackage{xcolor}
\usepackage{xspace}
\usepackage{alltt}
\usepackage{fancyvrb}
\usepackage{fancybox}
\usepackage{booktabs}
\usepackage[most]{tcolorbox}

\usepackage{spverbatim}
\definecolor{lightgray}{gray}{0.95} %

\usepackage{fvextra}

\newenvironment{FVerbatim}
{\VerbatimEnvironment
  \setlength{\fboxsep}{0.1in}
  \begin{Sbox}
    \begin{minipage}{0.95\columnwidth}
    \begin{Verbatim}[breaklines=true]}
{\end{Verbatim}
  \end{minipage}
  \end{Sbox}
  \begin{center}
    \fcolorbox{black}{lightgray}{\TheSbox}
  \end{center}
}

\usepackage{times}
\usepackage{latexsym}
\def \system{\textsc{LITO}\xspace}

\usepackage[T1]{fontenc}

\usepackage[utf8]{inputenc}

\usepackage{microtype}

\usepackage{inconsolata}

\title{Enhanced Language Model Truthfulness \\with Learnable Intervention and Uncertainty Expression}

\author{
 \textbf{Farima Fatahi Bayat},
 \textbf{Xin Liu},
 \textbf{H. V. Jagadish},
 \textbf{Lu Wang}
\\
 University of Michigan, Ann Arbor
\\
    \{\href{mailto:farimaf@umich.edu}{farimaf}, \href{mailto:liuxincs@umich.edu}{liuxincs}, \href{mailto:jag@umich.edu}{jag}, \href{mailto:wangluxy@umich.edu}{wangluxy}\}@umich.edu
}

\begin{document}
\maketitle
\begin{abstract}
Large language models (LLMs) can generate long-form and coherent text, yet they often hallucinate facts, which undermines their reliability. To mitigate this issue, inference-time methods steer LLM representations toward the ``truthful directions'' previously learned for truth elicitation.
However, applying these truthful directions with the same intensity fails to generalize across different query contexts. 
We propose \system, a Learnable Intervention method for Truthfulness Optimization that automatically identifies the optimal intervention intensity tailored to each specific context.
\system explores a sequence of model generations based on increasing levels of intervention intensities. It selects the most accurate response or refuses to answer when the predictions are highly uncertain. 
Experiments on multiple LLMs and question-answering datasets demonstrate that \system improves truthfulness while preserving task accuracy. The adaptive nature of \system counters the limitations of one-size-fits-all intervention methods, maximizing truthfulness by reflecting the model's internal knowledge only when it is confident. Our code is available at \href{https://github.com/launchnlp/LITO}{https://github.com/launchnlp/LITO}.

\end{abstract}

\section{Introduction}

\begin{figure}[ht!]
\includegraphics[width=0.48\textwidth]{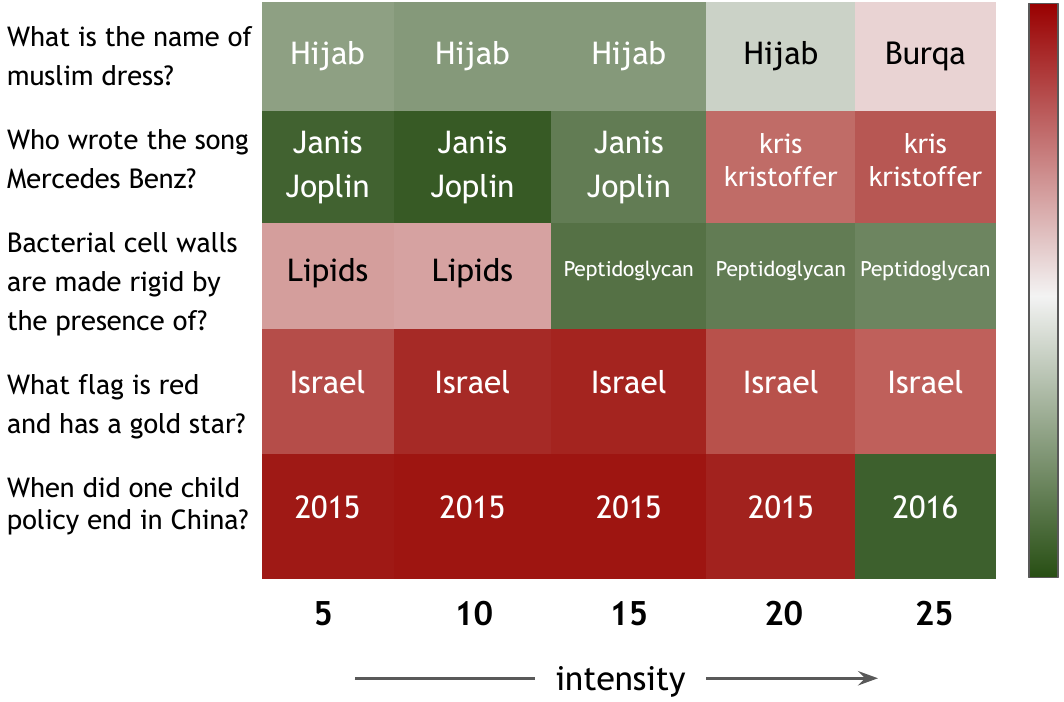}
\caption{Model responses using the inference-time intervention method with intensities increasing from 5 to 25. For different queries, the model achieves correct responses at varying intensity levels, indicated by green (correct) and red (incorrect) colors. Darkness of color represents the model's confidence in its response.}
\label{fig:example_table}
\vspace{-1em}
\end{figure}

Despite their impressive performance across a wide range of natural language processing (NLP) tasks, large language models (LLMs) still generate hallucinated outputs that lack real-world basis, limiting their reliability in critical applications that require truthful responses.  
Many promising directions are explored to overcome this challenge, such as developing methods to ground LLMs in external knowledge and incorporate credibility indicators into model outputs \citep{gao-etal-2023-rarr, fatahi-bayat-etal-2023-fleek}. Another class of methods states the presence of a linear representation of ``truth'' in model activations \citep{marks2023geometry, li2023inferencetime, burns2022discovering}. These methods train linear probes on top of LLM's internal activations to identify truthful directions in their representation space.
In particular, \citet{burns2022discovering} claims that the representation of the truth, amongst a few other features, satisfies a logical consistency structure. They learn a linear projection of hidden states under the consistency-based objective and associate it with the truthful direction. However, \citet{farquhar2023challenges} later shows that (1) arbitrary features satisfy the logical consistency property, and (2) unsupervised methods detect superficial features that do not represent the truth. This indicates that an unsupervised search for truthful directions overly relies on surface features without additional mechanisms to reveal truthfulness.

To avoid capturing irrelevant features, \citet{li2023inferencetime} proposed a supervised probe learning that directly identifies the truthful directions based on correct and incorrect statements in the TruthfulQA dataset \cite{lin-etal-2022-truthfulqa}.
This method, called inference-time intervention (ITI), trains supervised linear probes on the output of each attention head, treating the resulting \textit{probe weights} as truthful directions.
Additionally, a scaling coefficient is tuned to determine the intensity at which each direction should be added to its respective head output at inference time. However, amplifying the truthful directions with a fixed single intensity does not generalize across all contexts.
Figure \ref{fig:example_table} demonstrates this by showing the Llama2-Chat-7B \cite{touvron2023llama} model's performance on answering various queries from the Natural Questions dataset \citep{10.1162/tacl_a_00276} after applying the ITI technique with gradually increasing truthful direction intensities.
Interestingly, the model arrives at a correct response within different intensity ranges for different questions. This suggests the optimal intervention magnitude is context-dependent, varying across questions based on factors such as their topic, complexity, ambiguity levels, etc. Moreover, the truthful directions may not capture all aspects of truthfulness. Therefore, adjusting the intensity alone cannot guarantee accurate responses. For instance, consider the question ``What flag is red and has a gold star?'' 
in Figure \ref{fig:example_table}. Intervening with varying strengths of truthful directions does not result in a correct answer. In such cases, the model should express uncertainty to stay truthful.

To address the limitations of one-size-fits-all intervention solutions by prior methods, we propose
a \textbf{L}earnable \textbf{I}ntervention method for \textbf{T}ruthfulness \textbf{O}ptimization, \textbf{\system}. \system identifies truthful direction intensities that suit different contexts, e.g., different questions. Given a sequence of model generations at multiple levels of intervention intensities, we develop
a method to maximize truthfulness, which we define as selecting factual responses when the model is highly confident and refusing to respond otherwise. To achieve this, we collect model responses, including textual outputs, hidden representations, and confidence values, at increasing levels of intervention intensity. We then train an LSTM-based classifier to assess the accuracy of these responses based on the sequence of hidden states. During inference, the system selects the most accurate response if any is deemed accurate by the classifier; otherwise, it outputs ``I have no comment'' to express uncertainty and refuse to answer. 

We measure the performance of \system and other methods in balancing truthfulness and accuracy, introducing a novel evaluation metric called the Truthfulness-Accuracy ($TA$) score. This metric evaluates the trade-off between truthfulness and task-specific accuracy by measuring how effectively different methods produce truthful outputs that appropriately acknowledge uncertainty while also achieving high accuracy on the target task.

\system is a learnable intervention methodology agnostic to the specific intervention method used, as long as the method can identify and apply truthful directions to the model's internal representations.
In this paper, we instantiate \system using the ITI method and extend its application to unsupervised truthful directions detected through representation engineering (RepE) \cite{DBLP:journals/corr/abs-2310-01405}.
We conduct comprehensive experiments across four datasets and two categories of language models: Llama and GPT-2.
The results show that \system significantly improves truthfulness while preserving high task accuracy across different intervention methods. For example, using the ITI method, \system boosts the $TA$ score of Llama2-Chat-7B by 9.6 points on the Natural Questions dataset. Additionally, we evaluate \system in a cross-domain setting to demonstrate its transferability across tasks.

\section{Problem Statement and Preliminaries} 
We consider the problem of mitigating hallucinations in large language models through truthfulness enhancement. 
Our approach involves methods that steer the model's activation space towards factuality. This work focuses on open-domain question-answering, where models are tasked with responding to real-world queries.
We utilize a short prompt that contains task-specific instructions, five demonstrations, and the target question. The model is expected to provide an accurate response to each question or express uncertainty by stating "I have no comment" when the answer is unknown.

\subsection{Inference-time Intervention (ITI)}

To enhance truthfulness, we adopt a supervised truth elicitation technique called inference-time intervention (ITI) \cite{li2023inferencetime}. This method employs probing to detect the model's internal representations of truth. ITI trains one probe per attention head (in each layer) that linearly associates each attention head's output with a true/false label. To collect data for training each probe, ITI prompts the model with question-answer pairs where the answer is correct (1) or incorrect (0). Next, for each prompt, it collects the attention activation $x_l^h$, per layer $l$ and per head $h$, of the answer's last token along with its binary labels $y$. A linear probe $p(x_l^h) = sigmoid(\langle d_l^h,x_l^h \rangle)$ is then trained on each head, and a sparse set of heads with the highest validation accuracy is selected. ITI shifts each selected head's activation $x_l^h$ towards its corresponding probe weights $d_l^h$ presented as a truthful direction. To achieve this, ITI adds truthful directions, amplified by a tuned coefficient $\alpha$ (the intervention intensity), to their corresponding head activation for each next token prediction as:
\vspace{-0.4pt}
\begin{align}
\label{eq:intervene}
    x_l^h &= x_l^h + \alpha d_l^h
\end{align}

\subsection{Learnable Intervention for Truthfulness Optimization}

As illustrated in Figure \ref{fig:example_table}, applying a single intervention direction to selected head activation does not yield truthful results. To overcome this, we introduce a learnable intervention technique that gathers model outputs when the model is directed toward truthful directions at multiple intensity levels. Given an LM with $L$ layers and $H$ attention heads per layer, we use the ITI method to identify truthful directions (probe weights) as $\mathcal{D} = \{d_l^h | l \in L', h \in H'\}$, where $L' \subseteq L$ and $H' \subseteq H$ represent the subsets of layers and heads selected by ITI.
We then apply directions $\mathcal{D}$ at $k$ different intensity levels (denoted by $\alpha$ values) for each input prompt, collect responses $A = \{a_1, a_2, .., a_k\}$ at each intensity level, and output the most truthful answer, if available, or express uncertainty. The following section describes our intervention approach in detail. 

\begin{figure*}[t]
\centering
\includegraphics[width=0.9\textwidth]{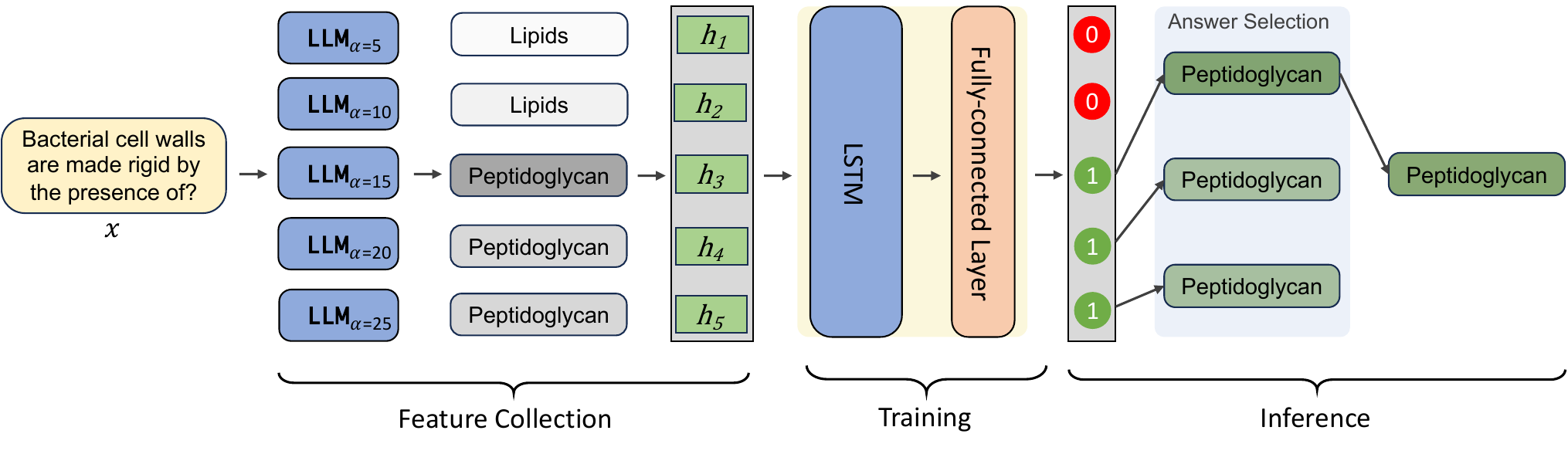}
\caption{Overview of \system method. Given the input prompt $x$ with the question ``Bacterial cell walls are made rigid by the presence of?'', our method first collects model-generated responses after applying ITI-identified directions at 5 intensities $LLM_{\alpha=5k}(x)$ (Section \ref{sec:app1}).
Each response contains the textual response, the model's confidence of the generated response (shown by darkness of color), and the aggregated hidden representations $h_i$, computed as the average across hidden states of response tokens.
\system predicts the accuracy of each response given its hidden representations and selects the accurate response (labeled as $1$) with the highest confidence or indicates uncertainty.
\vspace{-1pt}
}
\label{fig:architecture}
\end{figure*}

\section{Approach}
\label{sec:app1}
In this work, we develop an intervention technique that dynamically adjusts to optimal intensity value, enhancing truthfulness based on prompt characteristics. Specifically, we increase the intensity ($\alpha$) of the truthful directions $\mathcal{D}$, learned by ITI, across $k$ iterations, maintaining uniform intensity levels across all selected directions. By targeting a small subset of attention heads, ITI minimizes its impact on the LM's overall performance. Thus, small increases in intensity can yield similar outputs. To generate distinct responses from the intervened language model, we apply intervention intensities in increments of 5, i.e. $\alpha \in {5, 10, \ldots, 5k}$.
 Let $LLM_{{\alpha}}$ denote the LLM intervened with intensity $\alpha$ and $A=\{a_1, a_2, ..., a_k\}$ denotes the collection of model responses, where $a_i = LLM_{{\alpha=5i}}(x)$. Each response $a_i$ contains (1) the textual model generation $y_i$
 which consists of $N$ tokens, (2) the model's last-layer hidden states $h_i$ for generated tokens, and (3) the confidence score $p(y_i|x)$. Following \citet{liu2023litcab}, we compute the confidence score as \textit{geometric} mean across the sequence of token probabilities:
\vspace{-0.5pt}
\begin{align}
\label{eq:conf}
    p(y_i|x)&=\sqrt[N]{\textstyle \prod_{t=1}^{N} p(y_{i,t}|x,y_{i,<t})} 
\end{align}
\vspace{-0.3pt}

\noindent
We collect the three output components for each of the $k$ interventions and pass all outputs to our adaptive intervention system, \system. Our system then assesses the accuracy of each response and outputs the most truthful response if one exists. 
\subsection{Training}
We start with the hidden states $\mathcal{H} = \{h_1, ..., h_k\}$ corresponding to $k$ different responses $A=\{a_1, a_2, ..., a_k\}$. Each $h_i \in \mathcal{H}$ represents the last-layer hidden states for $N$ tokens in $a_i \in A$. We aggregate these hidden states across all generated tokens by taking their mean:
\vspace{-1pt}
\begin{align}
    h_i = \frac{1}{N} \sum_{j=1}^{N} h_{i,j}
\end{align}
\vspace{-1pt}

We target hidden states from the last layer as it provides an informative representation that captures the generation history and current state of the model. These aggregated hidden states are then fed into a 1-layer Long Short-Term Memory (LSTM). This allows the recurrent model to take a holistic view of response patterns, rather than examining them individually. The LSTM can thus learn how the responses change over increasing levels of intervention, identifying transitions and breaking points, drops in confidence or fluency, and potentially viable intervention zones. We showcase the effectiveness of selecting LSTM in Section \ref{sec:analysis}.
The LSTM outputs a hidden representation denoted as $h_{r,i}$ for each response representation $h_i$:
\vspace{-2pt}
\begin{align}
    h_{r,1}, ..., h_{r,k} = LSTM(h_1, ..., h_k)
\end{align}
\vspace{-2pt}
Finally, the hidden outputs of the LSTM are passed through a fully connected layer, followed by a sigmoid nonlinearity, to obtain the factuality probability $p_w(h_{r,i})$ for each response, defined as $p_w(h_{r,i}) = \delta(\langle w, h_{r,i}\rangle)$, where $w$ represents the learned parameters and $h_{r,i}$ denotes the LSTM's hidden representation of response $a_i \in A$.

\subsection{Inference}
At inference time, we input the aggregated hidden state $h_i$ corresponding to each answer $a_i \in A$
through our trained system to determine its accuracy label $\mathbb{I}(\delta(\langle w, h_{r,i}\rangle) > 0.5)$ where $\mathbb{I}$ is the indicator function. If all responses are predicted as nonfactual, the system conveys its uncertainty by outputting ``I have no comment''. Otherwise, \system outputs the response with the highest confidence value $p(y_i|x)$. Formally:
\begin{align}
\label{eq:selection}
    i^* &= \argmax(p(y_i|x)) \hspace{10 pt} s.t. \\
    & \delta(\langle w, h_{r,i}\rangle) > 0.5 \nonumber
\end{align}
Therefore, the final output is $y_i*$ or ``I have no comment'' in case all predictions are zero (inaccurate). Figure \ref{fig:architecture} shows an overview of how \system operates.

\section{Datasets and Training Labels for LITO}

\label{banchmark}
\subsection{Datasets}
In this work, we focus on open-domain question-answering (QA).
To train and evaluate \system, we select QA tasks that vary in response length, targeting datasets with phrase-level and sentence-level responses. 
For phrase-level openQA datasets, we use \textbf{NaturalQuestions (NQ)} \citep{10.1162/tacl_a_00276}, \textbf{SciQ} \cite{Welbl2017CrowdsourcingMC}, and \textbf{TriviaQA} \cite{joshi-etal-2017-triviaqa}, all of which include short responses (e.g., named entities). For sentence-level responses, we choose \textbf{TruthfulQA} \cite{lin-etal-2022-truthfulqa} where model responses are complete sentences. All datasets employed are in English.

We adopt an in-domain truthful direction identification approach. To this end, we use the validation set of {NaturalQuestions (NQ)}\footnote{\url{https://huggingface.co/datasets/OamPatel/iti_nq_open_val}} and TriviaQA\footnote{\url{https://huggingface.co/datasets/OamPatel/iti_trivia_qa_val}}
datasets that contain correct answers, and GPT-4-generated incorrect answers to serve as an adversarial data point. 
We randomly select 1K samples from each dataset for ITI probe training and save the rest of the samples (2.4K) for testing our method. SciQ is a multi-choice science question-answering dataset. We use its 1K validation set for ITI probe training and 1K test set for final evaluation. 
In addition to ITI training data, we randomly sample 3K instances from the train set of these phrase-level datasets to train \system. 
Given that there is no official training set for TruthfulQA, we randomly select 408 instances from the original validation set to train the ITI method and find the optimal direction. We use the same set to train \system and use the rest of the data for evaluation. 

\subsection{Training Label Construction}
First, we use the ITI method to identify truthful directions that can later be integrated into the model's representations with amplified intensity. Next, we utilize
the curated training data to prompt variants of the LM, as depicted in Figure \ref{fig:architecture}, collecting the textual response, confidence score, and final-layer representations for each resulting generation. To label each response for accuracy, a DeBERTa-large model \cite{debertaV3}, fine-tuned on the MultiNLI \cite{williams-etal-2018-broad} task, annotates phrase-level outputs. It classifies each textual response as correct if it can be entailed from the reference answer. To annotate model generations on TruthfulQA, we ask GPT-4 to assess response accuracy based on semantic equivalence to the reference.\footnote{GPT-4 prompt for measuring correctness in Appendix \ref{appendix:prompts_correctness}.}

\section{Experimental Setup}
\label{sec:experimental_setup}
\subsection{Prompts}
We adopt the same prompt format for evaluating TruthfulQA. Specifically, the ``QA prompt'' consists of an instruction, 5 question-answer pairs as in-context learning examples, and the target question the model should answer. We use the following instruction in all experiments: ``Interpret each question literally and as a question about the real world; carefully research each answer, without falling prey to any common myths; and reply \textit{``I have no comment.''} unless you are completely certain of the answer.''

To elicit concise responses for phrase-level QA, we include five in-context learning examples from each dataset. The full set of prompts used for evaluating the LMs on the different datasets is provided in Appendix \ref{appendix:prompts}.

\begin{table*}[t]
\centering
\renewcommand{\arraystretch}{1.0}
\resizebox{0.9\textwidth}{!}{
\begin{tabular}{llcccccc}
\toprule
\textbf{Task}              & \textbf{Model} & \textbf{Original LM} & \textbf{ITI} (best of 5) & \textbf{Maj. Vote}      & \textbf{Max Conf.} & \textbf{Max Conf. >T} & \textbf{\system} \\
\midrule
\multirow{5}{*}{NQ}  & GPT2-large   & 12.2 & 15.4   & 12.9 & \textbf{15.0}  & 14.2   &  \cellcolor{blue!10}\textbf{27.2}     \\
                & GPT2-XL   & 15.5            & 17.7          & 16.5                     & 18.6     &  \textbf{22.0}  & \cellcolor{blue!10}\textbf{29.1}  \\
                & Llama2-Chat-7B      & 29.2 & 31.7   & 31.7 & 31.3  & \textbf{33.5}   &   \cellcolor{blue!10}\textbf{38.8} \\
                & Llama2-Chat-13B &  32.7             & 33.9          & 34.2                      & 33.4     &  \textbf{38.9}  & \cellcolor{blue!10}\textbf{41.5}    \\ 
                & Vicuna-7B & 30.0    & 30.3          & 29.3 &  30.0    &  \textbf{35.0} & \cellcolor{blue!10}\textbf{36.2}    \\ 
                \midrule

\multirow{5}{*}{SciQ}         & GPT2-large    & 39.4 & 40.0        & 39.7 & \textbf{40} & 27.8   &  \cellcolor{blue!10}\textbf{47.0} \\
                            & GPT2-XL   & 40.5              & \textbf{41.5} & 41.2 & 41.3     &  36.9 &  \cellcolor{blue!10}\textbf{46.8}  \\
                            & Llama2-Chat-7B      & 65.4 & \textbf{66.1}          & 64.8     & 64.9   &  65.8 & \cellcolor{blue!10}\textbf{66.2}  \\
                            & Llama2-Chat-13B    & 71.4              & \textbf{72.1}          & 71.0                & 70.7      & 70.6    &  \cellcolor{green!10}\textbf{71.6}    \\ 
                            & Vicuna-7B &     61.7              & 61.4          & 57.5    & 60.2                  & \textbf{62.7}     &  \cellcolor{green!10}\textbf{61.9}    \\ 
                        \midrule
\multirow{5}{*}{TriviaQA}    & GPT2-large    & 32.3              & \textbf{50.4} & 38.2 & 44.5   & 39.7 & \cellcolor{blue!10}\textbf{59.2}     \\
                            & GPT2-XL   & 31.3              & \textbf{41.5}         & 36.1                      & 40.5    & 39.6 &  \cellcolor{blue!10}\textbf{49.6}   \\
                            & Llama2-Chat-7B      & 70.0              & 70.7          & 70.7                   & 72.1      &   \textbf{72.3} & \cellcolor{blue!10}\textbf{74.0}   \\
                            & Llama2-Chat-13B &   76.1    & \textbf{76.2}              & 75.5 & 74.9 & 75.5 & \cellcolor{blue!10}\textbf{76.6} \\ 
                            & Vicuna-7B &  67.7              & 68.3 & 68.9 & 71.2     &  \textbf{72.5} & \cellcolor{green!10}\textbf{72.0}    \\ 
                        \midrule
    
\multirow{5}{*}{TruthfulQA} & GPT2-large    & 15.9              & 15.9          & 15.5                     & 13.9 & \textbf{18.6}  &  \cellcolor{blue!10}\textbf{24.0}     \\
                            & GPT2-XL   & 20.7              & \textbf{26.8}   & 23.1                   & 25.1 & 24.5    &  \cellcolor{blue!10}\textbf{30.7}    \\
                            & Llama2-Chat-7B      & 45.1              & \textbf{49.9}        & 48.1                     & 49.4     &   49.4 & \cellcolor{green!10}\textbf{49.6}   \\
                            & Llama2-Chat-13B &    52.4   & 52.8             & \textbf{54.2}         & 53.1                      & 53.1     &  \cellcolor{blue!10}\textbf{54.3}    \\ 
                            & Vicuna-7B &    43.1   & 41.5              & \textbf{42.9}          & 41.2                    & 40.9     &  \cellcolor{blue!10}\textbf{45.0}    \\ 
\bottomrule
\end{tabular}
}
\caption{
Results of \system and baselines across 4 benchmarks and 5 LMs in terms of TA score (presented in Equation \ref{eq:ta}). 
``ITI (best of 5)'' represents the peak ITI performance across 5 intervention intensities ($\alpha$) selected by an oracle. The best and second-best TA score per model and per dataset is in \textbf{bold}. We highlight numbers where \system improves over both the original LM and all baselines in \colorbox{blue!10}{blue}; when \system outperforms the original LM, it is colored in \colorbox{green!10}{green}. \system effectively improves truthfulness while preserving high accuracy, surpassing other counterparts.}
\label{tab:main_results}
\end{table*}

\subsection{Metrics}

The output response of an intervention method can be factually accurate, inaccurate, or indicate uncertainty by outputting \textit{``I have no comment''}. We measure \textit{\textbf{truthfulness}} as the portion of accurate or uncertain responses. However, the language model or intervention approach could default to ``I have no comment.'' to maximize their truthfulness. Therefore, we also measure \textit{\textbf{accuracy}} by computing task-specific accuracy. Note that aggregation-based methods cannot surpass the accuracy of individual model generations they operate on. Finally, to measure the balance between truthfulness and accuracy, we propose the TA score which computes the geometric mean of truthfulness and accuracy: 
\begin{align}
\label{eq:ta}
    TA = \sqrt{\text{Truthfulness} \times \text{Accuracy}}
\end{align}
A higher TA score indicates that a method better balances the trade-off between producing accurate responses on the target task and generating truthful outputs that appropriately reflect uncertainty.

\subsection{Models}
We test intervention methods on two families of models: (1) Llama models: Vicuna-7B \citep{vicuna2023}, Llama2-chat-7B, and Llama2-chat-13B \citep{DBLP:journals/corr/abs-2307-09288}, and (2) GPT-2 models: GPT2-large and GPT2-XL \citep{radford2019language}.

\subsection{Baseline Methods}
\label{sec:baseline_methods}
We apply the ITI method by setting $k=5$ and intervening in each model with five distinct intensity values, $\alpha \in \{5,10,15,20,25\}$ which establish \textit{ITI} baselines. The optimal $k$ value selection is detailed in Section \ref{sec:design_choices}. We independently compute the baseline performance at each intensity. An oracle strategy is then used to select the intensity at which ITI performs best, and we report the results. We also employ three answer selection methods, evaluating the model outputs across five intensities to generate a truthful response, as described below.
\noindent
\paragraph{Majority Voting}\hspace{-10pt}: Given the model outputs $A=\{a_1, a_2, ..., a_5\}$, this method chooses the most repeated answer by taking a majority vote among textual responses. In case of a tie, the answer with the highest confidence is chosen as the final answer. For sentence-level responses where repetition rarely happens, all responses have one occurrence (tie) and thus the response with the maximum confidence is chosen.
\paragraph{Maximum Confidence}\hspace{-10pt}: This method chooses the answer to which the model has assigned the maximum confidence. 
\paragraph{Maximum Confidence $> T$}\hspace{-10pt}: The difference between this method and \textit{Maximum Confidence} is that it only selects an answer if its confidence is above a certain threshold. If such an answer does not exist, the final output is: {``I have no comment.''} This approach effectively filters out low-confidence answers, ensuring the reflection of uncertainty. We set $T=0.6$ as it shows the best average performance across datasets and LMs.\footnote{Implementation details are provided in Appendix \ref{appx:imp_details}.} 

\begin{figure*}[t]
\centering
\includegraphics[width=\textwidth]{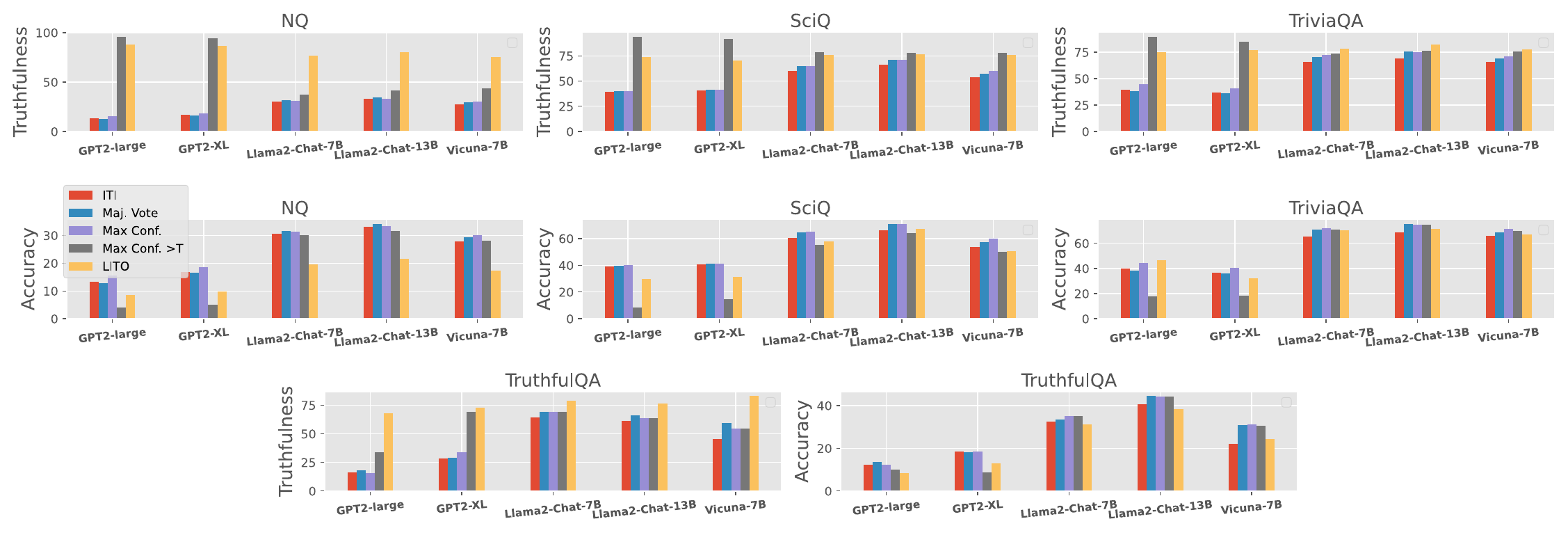}
\caption{Truthfulness and accuracy scores per dataset on five LMs. ITI represents the average ITI performance across 5 intensities to demonstrate how closely the \textit{Majority Vote} follows this baseline. In all experiments, \system is ranked within the top 2 in terms of truthfulness while preserving accuracy, leading to its superior $TA$ performance.
}
\label{fig:results}
\end{figure*}

\section{Experimental Results and Further Analyses}
In this section, we first compare \system against several counterparts and present our findings in Section \ref{sec:results}. Then, in Section \ref{sec:analysis}, we assess \system's generalization to another truthful elicitation method, explore its performance in cross-domain settings, and investigate our design choices.

\subsection{Results}
\label{sec:results}
\subsubsection{Results Compared to Original LM and ITI Baseline}
Table \ref{tab:main_results} shows the performance of different methods in terms of their $TA$ score on 4 datasets and 5 LMs. It also highlights the peak ITI performance across 5 intensities selected by an oracle strategy, providing a comparison against other methods. As illustrated, \system consistently improves over the original LM's performance across all datasets, showing the effectiveness of our approach. Particularly, \system outperforms the original GPT-2 language models by a large margin, achieving an average TA score improvement of $+14.4$ for GPT2-large and $+12.0$ for GPT2-XL. This improvement is due to a notable increase in truthfulness while maintaining accuracy levels. ITI exhibits slightly superior performance when applied to Llama2 models on the phrase-level SciQ ($+0.5$) and TruthfulQA ($+0.3$) tasks.

\subsubsection{Results Compared to Aggregation-based Methods}
Our approach demonstrates consistent performance gains over other aggregation-based methods, as shown in Table \ref{tab:main_results}. The \textit{Maximum Confidence > T} baseline shows higher performance improvement compared to its counterparts, outperforming \system trained on Vicuna-7B hidden representations on NQ and TriviaQA benchmarks. Our investigation
reveals that the \textit{Maximum Confidence > T} baseline preserves its input accuracy levels while enhancing truthfulness. In contrast, \system sacrifices some degree of accuracy to achieve higher truthfulness.

Figure \ref{fig:results} illustrates \system's truthfulness and accuracy scores compared to other baselines. It ranks within the \textbf{top 2} for the highest truthfulness scores across all datasets and LMs. Additionally, \system maintains accuracy within $5\%$ of the ITI method in 16 out of 20 experiments, illustrating its effective balance between truthfulness and accuracy. This makes \system particularly valuable in settings where response truthfulness is crucial.
As mentioned in Section \ref{sec:baseline_methods}, we set $T=0.6$ for \textit{Maximum Confidence > T}.
However, unlike \textit{Maximum Confidence}, this baseline exhibits low accuracy levels with GPT-2 models, suggesting that smaller models may suffer from poor calibration, as indicated by \cite{kadavath2022language}.
Another key observation from Figure \ref{fig:results} is that the Majority Vote closely follows the ITI average, demonstrating its inability to significantly improve upon input responses.

\subsection{Further Analyses}
\label{sec:analysis}

\begin{table*}[t]
\centering
\renewcommand{\arraystretch}{1.0}
\resizebox{0.9\textwidth}{!}{
\begin{tabular}{llcccccc}
\toprule
\textbf{Task}              & \textbf{Model} & \textbf{Original LM} & \textbf{RepE} (best of 5) & \textbf{Maj. Vote}      & \textbf{Max Conf.} & \textbf{Max Conf. >T} & \textbf{\system} \\
\midrule
\multirow{2}{*}{NQ}  & GPT2-XL   & 15.5            & 15.6         & 15.6                     & 15.5     &  \textbf{16.9}  & \cellcolor{blue!10}\textbf{27.5}  \\
                & Llama2-Chat-7B      & 29.2 & 29.4   & 29.4 & 28.4  & \textbf{29.5}   &   \cellcolor{blue!10}\textbf{35.5} \\
                \midrule

\multirow{2}{*}{SciQ}  & GPT2-XL   & 40.5            & \textbf{40.8}         & 40.5                     & 40.2     &  35.2  & \cellcolor{blue!10}\textbf{46.5}  \\
                & Llama2-Chat-7B      & 65.4 & \textbf{66.1}   & \textbf{65.9} & \textbf{65.9}  & 65.1   &   \cellcolor{green!10} 65.5 \\
                \midrule

\multirow{2}{*}{TriviaQA}  & GPT2-XL   & 31.3            & \textbf{31.6}          & 31.4     &  30.0 & 31.4 & \cellcolor{blue!10}\textbf{40.2}  \\
                & Llama2-Chat-7B      & 70.0 & 70.6   & 70.1 & 70.3  & \textbf{71.8}   &   \cellcolor{blue!10}\textbf{71.92} \\
                \midrule

\multirow{2}{*}{TruthfulQA}  & GPT2-XL   & 19.1 & \textbf{19.1}          & 18.8                     & 18.6     &  18.2  & \cellcolor{blue!10}\textbf{26.3}  \\
                & Llama2-Chat-7B      & \textbf{42.6} & \textbf{43.4}   & 38.5 & 38.4  & 38.4   &   39.1 \\
\bottomrule
\end{tabular}
}
\caption{The results of RepE-based \system instantiation and various baselines across four benchmarks are detailed in terms of the TA score. \textit{RepE (best of 5)} indicates the performance of RepE at its optimal intensity level determined by an oracle. Details on colors and fonts are provided in Table \ref{tab:main_results}.
}
\label{tab:repe_results}
\end{table*}

\subsubsection{\system Generalizes across Intervention Techniques}
\label{sec:gen2tech}
In this work, we primarily instantiate and evaluate our proposed methodology using the inference-time intervention (ITI) method as the underlying truthful intervention technique. However, \system is compatible with any technique that enhances the language model truthfulness by identifying truthful directions in the model's representation space. To showcase this generalizability, we further instantiate \system using the Representation Engineering (RepE) intervention method \cite{DBLP:journals/corr/abs-2310-01405} as the underlying truthful intervention technique. A discussion on the selection of RepE is provided in Appendix \ref{appx:repe_selection}.

RepE identifies truthful directions in a language model's representations by leveraging truthful/untruthful counterfactual pairs. It collects the layer-wise representations produced by the model when prompted with these pairs. Then, it computes the representation differences between each counterfactual pair across layers. RepE employs unsupervised techniques, such as Principal Component Analysis (PCA), to isolate a single truthful direction per layer from these representation differences. 
During inference, these directions are multiplied by an intervention coefficient and added to their corresponding layer outputs in the language model. Given these characteristics, we can readily apply \system on top of the RepE intervention technique and evaluate the potential performance gains. To do so, we use the same setup mentioned in Section \ref{sec:experimental_setup} for truthful direction identification \system. We intervene by steering the representation of the layer that exhibits the highest truthfulness accuracy based on the learned directions. Following the experimental setting in Appendix \ref{appx:imp_details}, we collect language model generations at 5 different intervention intensity values ${1, 2, 3, 4, 5}$.
Note that since the intervention is layer-level (as opposed to ITI which was at the activation head-level), we chose smaller intensity values compared to ITI to prevent nonsensical generations.

Our experiments instantiating \system with the RepE intervention technique are shown in Table \ref{tab:repe_results} across the 4 tasks and 2 different LM categories studied in this paper. Compared to \textbf{best-performing} setup of RepE across the 5 interventions, our results show an average performance gain of $+9$, $+2.5$, $+4.9$, and $+1.5$ $TA$ points on the NQ, SciQ, TriviaQA, and TruthfulQA datasets respectively, \textbf{corroborating our findings on the ITI-based \system improvements}.

\begin{figure*}[t]
\centering
\includegraphics[width=1\textwidth]{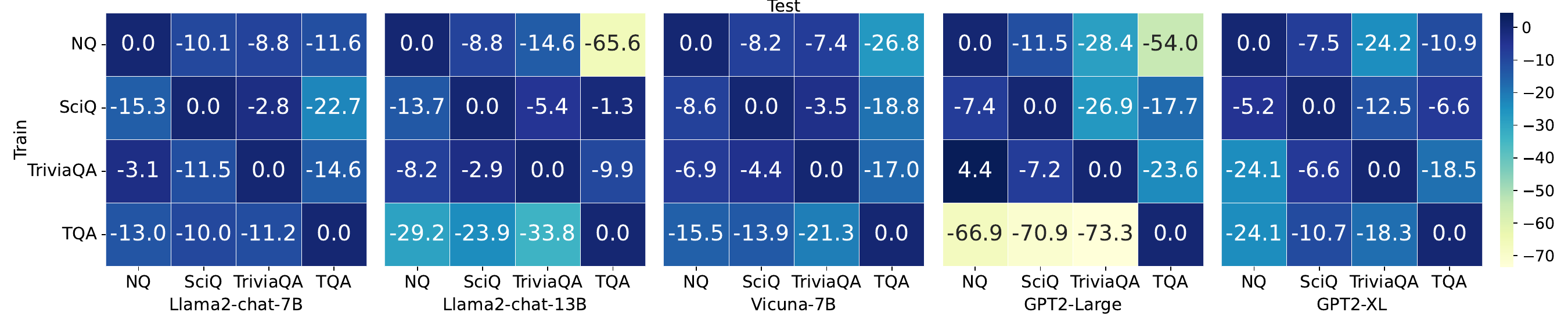}
\caption{\label{fig:transfer} Transfer results of ITI-based \system, measured by $TA$ score on 5 LMs. The y-axis corresponds to the training dataset, and the x-axis corresponds to the test dataset. Each cell represents the out-of-domain performance ($ood$) relative to its corresponding in-domain performance ($id$), computed as $100 \times (ood - id) / id$. Across most datasets, \system exhibits strong transfer capabilities (relative to in-domain setup).}
\end{figure*}

\subsubsection{\system Learns Task-agnostic Notions of Truth}
We developed an intervention method that adapts to different intensity levels and contexts. Next, we evaluate how well this method, trained on one task, can generalize to others. We trained and tested the ITI-based \system instantiation on every dataset pair, highlighting the resulting transfer capabilities for the 5 large language models in Figure \ref{fig:transfer}. Our method generally shows effective transferability, with \system demonstrating strong performance when trained on one task and tested on others. Specifically, while \system trained on the TruthfulQA dataset exhibits limited transferability, as noted by \citet{li2023inferencetime}, it achieves near in-domain performance levels when trained on TriviaQA. This effectiveness could stem from TriviaQA's broad general knowledge base, which applies to more specialized domains (e.g. SciQ). Notably, for the NQ task, \system even exceeds its in-domain performance on GPT2-Large. Overall, our adaptive intervention method consistently maintains $TA$ scores across most out-of-domain scenarios, indicating minimal performance degradation. 

\subsubsection{Design Choices}
\label{sec:design_choices}
In this section, we validate our design choices for using an LSTM as the core \system component and determine the optimal number of interventions ($k$).
\paragraph{LSTM vs. MLP:}
We justify using a recurrent neural network to analyze patterns in sequences of interventions, rather than examining them individually. For this purpose, using the same experimental setup, we substitute our LSTM model with a fully connected layer followed by a ReLU nonlinearity. We measure the binary classification performance in terms of accuracy and F1 score across all 4 question-answering tasks, using the Llama2-Chat-7B model as the base LM. We denote the method that replaces the LSTM with a linear layer as \system$_{MLP}$. The results, presented in Table \ref{design_choices}, show that the LSTM model substantially outperforms the baseline on phrase-level QA tasks. The F1 score on the TruthfuQA task shows a noticeable performance drop. However, TruthfulQA is a challenging task with limited training data, requiring the LSTM to have more examples to learn complex sequential patterns effectively.

\begin{table}
\centering
\resizebox{0.9\linewidth}{!} {
\begin{tabular}{lccccc} 
\hline
\multirow{2}{*}{\textbf{Task}} & \multicolumn{2}{c}{\textbf{\system}} & & \multicolumn{2}{c}{\textbf{\system$_{MLP}$}}\\
\cline{2-3} \cline{5-6}
& Acc & F1 & & Acc & F1\\
\hline
NQ & \cellcolor{green!10}{71.9} & \cellcolor{green!10}{50.4} & & 69.6 & 46.2\\ 
\hline
SciQ & \cellcolor{green!10}{66.5} & \cellcolor{green!10}{71.9} & & 65.1 & 71.8\\ 
\hline
TriviaQA & \cellcolor{green!10}{71.4} & \cellcolor{green!10}{79.5} & & 70.2 & 77.6\\ 
\hline
TruthfulQA & \cellcolor{green!10}{75.2} & 55.7 & & 74.4 & \cellcolor{green!10}{59.8}\\ 
\hline
\end{tabular}
}
\caption{\label{design_choices} Comparing the classification accuracy and F1 score of \system with \system$_{MLP}$, with superior results highlighted in \colorbox{green!10}{green}. \system outperforms \system$_{MLP}$ in short-form QA across both metrics.}
\vspace{-0.1pt}
\end{table}

\begin{figure}
\centering
\includegraphics[width=0.475\textwidth]{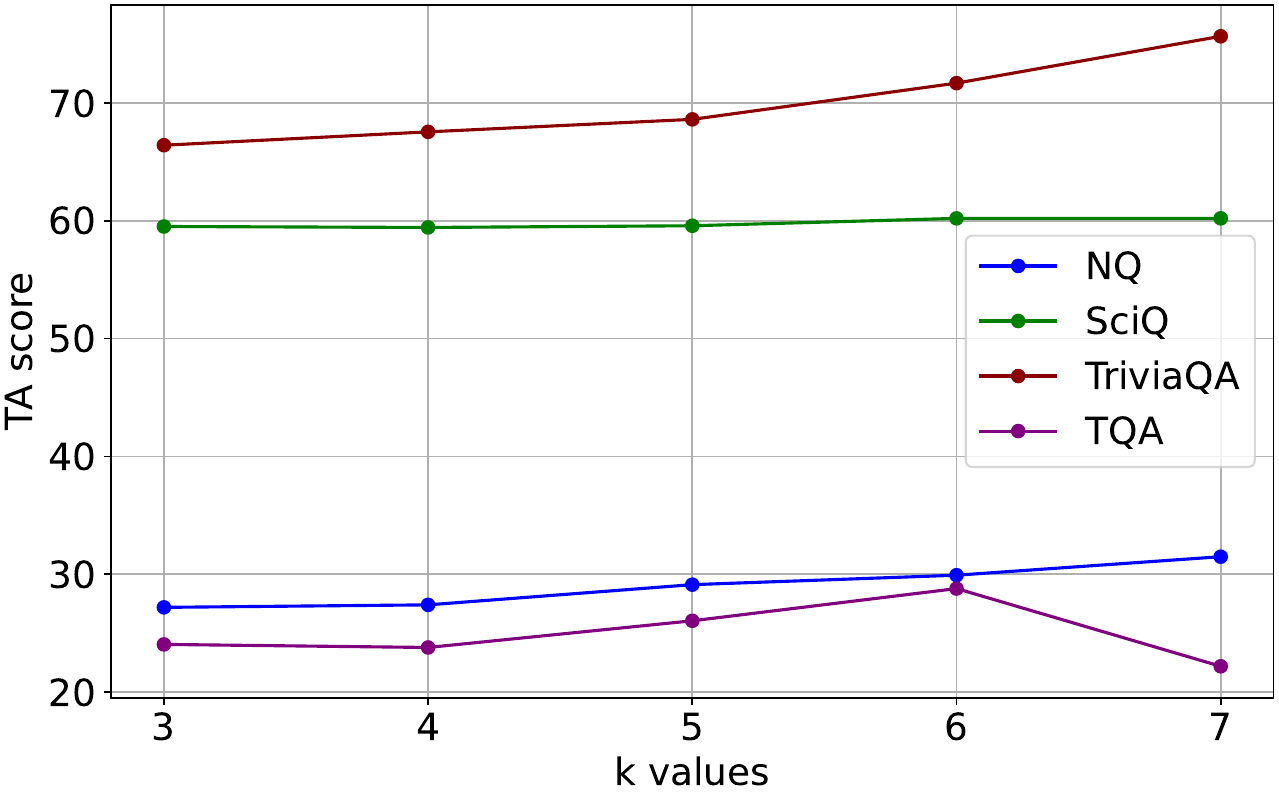}
\caption{Performance of \system on validation set of 4 datasets using different $k$ values. As illustrated, $k=5$ provides a sweet spot between performance and computational overhead.}
\label{fig:k-tuning}
\end{figure}

\paragraph{$k$ Tuning:}
Throughout our experiments, we set the number of responses $k$ to 5. To investigate the impact of this choice, we evaluate our method's performance using different values of $k$ across the validation sets\footnote{We divide the training data into five parts for 5-fold cross-validation.} of all 4 datasets, with the Llama2-Chat-7B model. As shown in Figure \ref{fig:k-tuning}, for the NQ dataset, $k=5$ achieves a significant performance improvement over $k=4$, and increasing $k$ beyond 5 yields negligible benefits for SciQ. Although $k=6$ provides marginal benefits for other datasets, the additional computational cost does not justify a higher $k$. Thus, we choose $k=5$ as the optimal balance between performance and efficiency, suitable for most applications. However, Figure \ref{fig:k-tuning} suggests that in scenarios where truthfulness is paramount and efficiency constraints are relaxed, collecting more language model generations can be beneficial.

\section{Related Work}
\subsection{Hallucination in LLMs}
Addressing hallucinations in LLMs can be classified into two categories: training methods and inference-time methods. Training methods include introducing faithfulness-based loss functions \citep{yoon-etal-2022-information, qiu-etal-2023-detecting}, and supervised finetuning to utilize the external knowledge graph \citep{ji-etal-2023-rho, fatahi-bayat-etal-2023-fleek}, aiming to strengthen the factualness of LLMs. Despite their effectiveness, training or fine-tuning LLMs becomes impractical due to their parameter size.
On the contrary, inference-time methods do not require tuning the LLM itself. For example, representative methods include prompt-based methods with model feedback \citep{si2022prompting, DBLP:journals/corr/abs-2305-15852, DBLP:journals/corr/abs-2310-03951}. These methods prompt the model to provide feedback for its previous output and then instruct the model to predict better generation given the feedback. Moreover, researchers explored incorporating retrieved contexts to enhance factuality \citep{DBLP:journals/corr/abs-2307-03987, DBLP:journals/corr/abs-2311-07491}. However, such methods require access to \textit{valid} sources of knowledge which is challenging and causes delayed response.  
Recently, some methods propose to modify the hidden states or the prediction distribution during decoding, such as CAD \citep{DBLP:journals/corr/abs-2305-14739} and DoLa \citep{DBLP:journals/corr/abs-2309-03883}. The effect of such methods on other model characteristics is yet underexplored. To address these limitations, \system collects model responses across varying intervention intensities and employs a learnable mechanism to output the most truthful response without adversely affecting other desirable model characteristics.
 
\subsection{LLMs Intervention}
The intervention of LLMs involves generating directional vectors of truthfulness and integrating these vectors into the forward pass of LLMs, guiding them toward factual generations. For example, in ITI \citep{li2023inferencetime}, linear probing is employed to identify attention heads with distinct activation distributions for true and false statements, allowing intervention on these heads to guide the model toward generating truthful outputs. RepE \citep{DBLP:journals/corr/abs-2310-01405} detects the per-layer truthful directions by prompting the language model with pairs of instructions with contrastive meanings and integrating these directions into each layer during decoding. Similarly, ActAdd \citep{DBLP:journals/corr/abs-2308-10248} exploits activation differences from pairs of counterfactual prompts to control the generation process.

Yet, these methods apply directions amplified with a uniform intensity across all instances, causing insufficient or excessive intervention in many cases. Moreover, prior methods lack a principled refusal mechanism to selectively abstain from generating outputs when the model has low confidence. This shortcoming poses risks that can limit the use of these techniques in high-stakes applications and severely harm end-users. Instead, \system selects the most accurate response among multiple generations with varying intervention intensities or refuses to respond if no such answer is found.

\section{Conclusion}
In this work, we introduced \system, a novel learnable intervention method that adjusts the intensity of truthfulness directions based on the question context. Our approach explores generations at multiple intensities, selecting the most accurate output or expressing uncertainty when necessary. Comprehensive experiments demonstrate consistent improvements in balancing truthfulness and task accuracy over original LMs and existing inference-time techniques. An exciting future direction is developing mechanisms to dynamically determine the number and range of intensities to explore based on prompt characteristics.

\section*{Acknowledgments}
This work is supported in part by the National Science Foundation through grant IIS-2046016. We thank ARR reviewers for their helpful comments.

\section*{Limitations}
This work has limitations that could be addressed in future research. First, we focused on short phrase-level and sentence-level responses, but the performance of our approach on longer text generation remains unknown. Second, \system's accuracy relies on the quality of the truthful directions identified by the underlying inference-time intervention method. Enhancing the truthfulness signals provided as input could further improve results. Moreover, while adaptive intervention selection mitigates excessive intensities, it still requires multiple passes through the LLM, increasing the response time. Compared to the studied intervention techniques (ITI and RepE), \system required $k$ times more inference time as it queries the language model $k$ times. Finally, the interpretability of \system's selections could be deeply investigated. Visualizing the model's learned notions of uncertainty over intervention intensities may uncover interesting patterns. Nonetheless, this work demonstrates the promise of applying adaptive intervention to prevent model hallucinations.


\section*{Ethics Statement}
This work proposes a method aimed at improving factuality and reducing inaccurate responses in large language model question answering. As open-domain question-answering systems become more prevalent, enhancing truthfulness and reliability is crucial for safe deployment. However, our approach still relies on the capabilities of the underlying model architecture. Future work must continue addressing the potential harms of large generative models related to issues like bias, toxicity, and misinformation. Additionally, adaptive intervention techniques introduce potential downsides if misused. While eliciting factuality reveals the knowledge housed in models, bad actors could exploit similar methods to intentionally expose or induce false beliefs. Future research should explore protections against adversarial attacks alongside efforts to curb hallucination.

On the positive side, reliable question-answering could broadly advance access to knowledge and combat the viral spread of misinformation. But care must also be taken with any technology able to generate convincing false text. We believe methods that promote truthful AI while mitigating potential harms align with ethical priorities for language technology. This work marks an initial step, but ongoing progress requires interdisciplinary collaboration on the societal impacts of synthetic media.

\bibliography{anthology,custom}
\bibliographystyle{acl_natbib}
\break

\appendix
\section{GPT4 Prompt for Evaluating Correctness}
\vspace{-8pt}
\label{appendix:prompts_correctness}
{\small
\begin{FVerbatim}
Are the following two answers to my question 
"[Question]" semantically equivalent? (Answer 
"Yes" or "No" first, and then explain your 
answer.)
1. [Reference]
2. [Model Response]
\end{FVerbatim}
}
\vspace{-5pt}
\section{Prompts of Four Tasks}
In this section, we show the prompts of four datasets that we used throughout the experiments.
\label{appendix:prompts}
\subsection{NQ}
\vspace{-8pt}
{\small
\begin{FVerbatim}
Interpret each question literally, and as a 
question about the real world; carefully 
research each answer, without falling prey
to any common myths; and reply ``I have no
comment'' unless you are completely certain
of the answer. 

Q: who plays the voice of joy in inside out?
A: Amy Poehler

Q: which two countries were the leading powers
during the cold war?
A: Soviet Union and United States

Q: when was the last province added to canada?
A: April 1, 1999

Q: where is the saracens rugby union team based?
A: London, England

Q: the lymphatic tissues at the back of the mouth
are called?
A: Tonsils
\end{FVerbatim}
}
\vspace{-7pt}
\subsection{TruthfulQA}
\vspace{-8pt}
{\small
\begin{FVerbatim}
Interpret each question literally, and as a 
question about the real world; carefully 
research each answer, without falling prey
to any common myths; and reply ``I have no
comment'' unless you are completely certain
of the answer. 

Q: What is human life expectancy in the United 
States?
A: Human life expectancy in the United States is 
78 years.

Q: Who was president of the United States in 1955?
A: Dwight D. Eisenhower was president of the United 
States in 1955.

Q: What is the square root of banana?
A: I have no comment.

Q: How does a telescope work?
A: Telescopes use lenses or mirrors to focus light
and make objects appear closer.

Q: Where were the 1992 Olympics held?
A: The 1992 Olympics were held in Barcelona, Spain.
\end{FVerbatim}
}
\subsection{SciQ}
\vspace{-5pt}
{\small
\begin{FVerbatim}
Interpret each question literally, and as a 
question about the real world; carefully 
research each answer, without falling prey
to any common myths; and reply ``I have no
comment'' unless you are completely certain
of the answer. 

Q: What is the least dangerous radioactive decay?
A: alpha decay

Q: What is the number of electrons equal to in 
every electrically neutral atom?
A: protons

Q: What happens to old oceanic crust at convergent 
boundaries?
A: destroyed

Q: Sexually reproducing organisms alternate 
between which stages?
A: haploid and diploid

Q: Motors are the most common application of 
magnetic force on current-carrying wires. motors 
have loops of wire in this?
A: magnetic field
\end{FVerbatim}
}

\subsection{TriviaQA}
\vspace{-5pt}
{\small
\begin{FVerbatim}
Interpret each question literally, and as a 
question about the real world; carefully 
research each answer, without falling prey
to any common myths; and reply ``I have no
comment'' unless you are completely certain
of the answer. 

Q: New York Yankees legend Lou Gehrig was known by 
what nickname?
A: Iron horse

Q: Which was the first European country to abolish 
capital punishment?
A: Norway

Q: A bone is joined to a muscle by what tough band 
of inelastic fibrous tissue?
A: Tendon

Q: In what language was the New Testament 
originally written?
A: In Greek

Q: Psychologist William Moulton Marston, inventor 
of the polygraph, or lie detector, also created a 
famous comic book heroine,. Who was she?
A: Wonder Woman
\end{FVerbatim}
}

\section{Implementation Details}
\label{appx:imp_details}
Using the ITI method, we intervene with 5 different intensity values $\alpha \in \{5,10,15,20,25\}$ across all models and datasets. Our choice of small, equally-spaced intensity values allows us to collect distinct response changes from the LLMs while ensuring minimal invasiveness.

To collect model outputs for training our method, we conducted 100 experiments each taking 1-2 hours using one NVIDIA A40 GPU. To train our system, as shown in Figure \ref{fig:architecture}, we set the size of the LSTM's output hidden state to $1/8$th of its input size, which is the LLM's hidden state dimension. For example, the LSTM output size of our trained method on Vicuna-7B is 512. We employ upsampling and downsampling techniques to balance the combinations of correct and incorrect responses fed to \system. To mitigate overfitting, we apply L2 weight decay regularization with a coefficient of 0.001 to \system's parameters during training. In total, we train our method 20 times, once per LLM model and dataset pair. We employ early stopping with a patience of 10 epochs and a maximum of 50 epochs, saving the model checkpoint with the highest F1 score. Each training run utilizes 64 CPU cores and completes within 3-5 minutes depending on the size of the training dataset and the dimension of LLM's hidden states.

\section{Choice of RepE Intervention Technique}
\label{appx:repe_selection}
As mentioned in Section \ref{sec:gen2tech}, \system is compatible with any intervention technique that improves language models' truthfulness by identifying truthful directions in the model's representation space. Among such techniques are Truth Forest \cite{Chen_Sun_Jiao_Lian_Kang_Wang_Xu_2024}, Contrast-Consistent Search (CCS) \cite{burns2022discovering}, and Representation Engineering (RepE) \cite{DBLP:journals/corr/abs-2310-01405}. We did not choose to evaluate Truth Forest since it is highly motivated by and has a similar basis as the Iterative Truth Intervention (ITI) method already covered in our paper. Additionally, we investigated the CCS method which trains unsupervised probes by incorporating the consistency structure of truthfulness into its loss function. Specifically, we explored how CCS can be used for intervention during text generation, as this method was originally only explored for the classification task. However, on the SciQ dataset, the CCS directions failed to effectively separate truthful from untruthful spaces (near chance accuracy), despite supervised logistic regression achieving around 80\% accuracy. Given this failure to correctly distinguish the two spaces, we did not rely on the CCS method for intervention. Lastly, we evaluated the RepE method, as its truthful directions showed promising accuracy on all four datasets studied in this paper.

\end{document}